\documentclass{article}
\usepackage{spconf,amsmath,graphicx}
\usepackage[tight,footnotesize]{subfigure}
\usepackage{bm}
\usepackage{amssymb}
\usepackage{enumitem}
\usepackage{algorithmic}
\usepackage{algorithm}
\usepackage{float,url}

\title{PAINTING ANALYSIS USING WAVELETS AND PROBABILISTIC TOPIC MODELS}
\name{Tong Wu, Gungor Polatkan, David Steel, William Brown, Ingrid Daubechies and Robert Calderbank 
\thanks{T. W. (\emph{tong.wu.ee@rutgers.edu}) is with the Department of Electrical and Computer Engineering, Rutgers University.
G. P. (\emph{gungorpolatkan@gmail.com}) is with Twitter Inc.
D. S. and W. B. (\emph{david.steel, william.p.brown@ncdcr.gov}) are with the North Carolina Museum of Art.
I. D. (\emph{ingrid@math.duke.edu}) is with the Department of Mathematics, Duke University.
R. C. (\emph{robert.calderbank@duke.edu}) is with the Department of Computer Science, Duke University.
T. W. performed the work while at Duke University. 
This work was supported in part by the Department of Homeland Security under grant HSHQDC-11-C. 
}
}
\address{}
\begin{document}
%
\maketitle
\begin{abstract}
In this paper, computer-based techniques for stylistic analysis of paintings  are applied to the five panels of the 14th century Peruzzi Altarpiece by Giotto di Bondone. Features are extracted by combining a dual-tree complex wavelet transform with a hidden Markov tree (HMT) model. Hierarchical clustering is
used to identify stylistic keywords in image patches, and keyword frequencies are calculated for sub-images that each contains many patches. A generative hierarchical Bayesian model learns stylistic patterns of keywords; these patterns are then used to characterize the styles of the sub-images; this in turn, permits to discriminate between paintings. Results suggest that such unsupervised probabilistic topic models can be useful to distill characteristic elements of style.
\end{abstract}
\section{Introduction}
\label{sec:intro}

In recent years wavelet methods have contributed to art history through their application 
to forgery detection \cite{PolatkanJBHD.ICIP2009}, linking of underdrawing and 
overpainting \cite{WolffMJDC.ICASSP2011}, and uncovering elements of style 
\cite{JafarpourPBHBD.EUSIPCO2009, LiYHW.PAMI2012}. 
The wavelet transform is a powerful tool which captures local features of an image at 
different resolutions; the characteristics of wavelet coefficients at different scales 
can be analyzed via hidden Markov tree models. When applied to brushwork, the combination 
of these feature extraction methods and machine learning algorithms is able to separate 
high-resolution digital scans of paintings by Vincent van Gogh from paintings by other artists 
(see Johnson et al. \cite{JohnsonHBBHDLPW.ISPM2008}).  Given the features extracted from the 
wavelet coefficients of the scanned images, several different classifiers can be usefully 
applied (see \cite{WolffMJDC.ICASSP2011, JafarpourPBHBD.EUSIPCO2009} for examples).

In this paper we seek to uncover the intrinsic styles in a dataset provided by the North Carolina 
Museum of Art, consisting of high-resolution scans of five panels attributed to Giotto di Bondone. 
In contrast to prior work \cite{WolffMJDC.ICASSP2011}, we lack side information such as 
underdrawings of different styles. As in \cite{PolatkanJBHD.ICIP2009, WolffMJDC.ICASSP2011}, we extract features from the dataset by 
combining a dual-tree complex wavelet transform \cite{SelesnickBK.ISPM2005} with hidden Markov trees
 \cite{ChoiRBK.ICASSP2000}. In the absence of a training set (such as a set of overpainting patches 
labeled by underdrawing styles in \cite{WolffMJDC.ICASSP2011}), we cannot apply supervised learning 
algorithms. Our main contribution is to use probabilistic topic models to model different 
styles present in each painting, where a style corresponds to a {\em topic} in the {\em bag of words} model. 
This produces a meaningful discrimination between the five paintings based on the style distribution 
in each painting.

This paper is organized as follows. Section \ref{sec:feature} reviews some background of wavelet-based 
HMT models applied to feature representation. Section \ref{sec:style} introduces our post analysis 
strategy via topic models. Results and discussion are provided in Section \ref{sec:result}; conclusions and future work are
in Section \ref{sec:con}.

\section{Feature Extraction}
\label{sec:feature}

\subsection{Color Representation}
\label{ssec:color}

We first introduce the HSL color representation, commonly used in computer graphics, and easily computed from RGB; 
it gives a perceptually more accurate representation of color relationships, superior to RGB. 
HSL \cite{Agoston.2004} describes colors by their \emph{hue} 
(coded by an angle on the color wheel, from 0 to 360), \emph{saturation} (ranging from 0 to 1) and 
\emph{lightness} (again from 0 to 1); this color space is often represented as a 
double-cone. This double-cone can be expressed in Cartesian coordinates (XYZ), 
computed from HSL values by setting
\noindent\begin{eqnarray}\label{eq:color}
    X &=& L  \nonumber \\
    Y &=& S \cos \left( ⁡ 2 \pi H / 360   \right)\, \min \{ 2L, 2(1-L) \} \\
    Z &=& S \sin \left( ⁡ 2 \pi H / 360   \right)\, \min \{ 2L, 2(1-L) \} \nonumber
\end{eqnarray}

\subsection{Wavelet Transforms}
\label{ssec:wavelet}

Wavelet transforms 
allow us to analyze images 
at different resolutions. By their finite support, wavelets are able to capture information 
specific to a particular location. Multiresolution analysis then provides information specific 
to both location and resolution or scale. The clustering and persistence properties of wavelet 
transforms provide us with statistical models to characterize the dependencies between wavelet 
coefficients \cite{CrouseNB.ITSP1998}; see also Section \ref{ssec:HMT}.

In the dual-tree complex wavelet transform (CWT) \cite{SelesnickBK.ISPM2005},
two separate discrete wavelet transforms (DWT) decompose signals into real and imaginary parts. 
The magnitude of each complex wavelet coefficient is nearly shift invariant, 
rendering CWT insensitive to small image shifts. The dual-tree CWT has six subbands 
and provides greater orientation selectivity than DWT. It represents local differences 
in terms of six basic directions, making it possible to capture brushstroke information 
specific to location, scale and orientation.

\subsection{Hidden Markov Trees}
\label{ssec:HMT}

The clustering and persistence properties 
of wavelet coefficients suggest the use of an HMT model to capture the intrinsic statistical 
structure of wavelet coefficients of an image; this provides an 
efficient dimensionality reduction technique that is robust to noise
induced by the scanning process \cite{ChoiRBK.ICASSP2000}.

At each level, the wavelet coefficients are modeled as a 
mixture of two Gaussian distributions: 
one, with large variance, corresponds to the wavelet component 
with high energy; the other, with small variance, corresponds to smooth components. 
Each HMT model thus has three parameters:
\begin{itemize}
\setlength{\itemsep}{0pt}%
\setlength{\parskip}{0pt}
  \item $\epsilon^{T}$, state transition probability between successive scales
  \item $\sigma^{S}$, variance of the narrow Gaussian distribution
  \item $\sigma^{L}$, variance of the wide Gaussian distribution
\end{itemize}

\algsetup{indent=0em}
\begin{algorithm}[t]
\caption{Feature Extraction}
\label{algo:featureextract}
\textbf{Input}: A set of painting images $M_i$, $ i\in \{1,2,3,4,5\}$.\\
\textbf{Initialize}: Divide images into $64 \times 64$ patches with $32$ pixels overlap, both vertically and horizontally. We term each patch $\{P_{i,j}\}_{i,j=1}^{5,n_i}$, where $n_i$ is the number of patches in image $M_i$.
\vspace*{1 mm}

\begin{algorithmic}[1]
\FOR {each patch $P_{i,j}$}
\STATE Convert $P_{i,j}$ into XYZ domain; normalize each domain.
\STATE Calculate complex wavelet coefficients $w_{i,j}^{x},w_{i,j}^{y},w_{i,j}^{z}$ and norms $w_{i,j}=\sqrt{ | w_{i,j}^{x} |^2+| w_{i,j}^{y} |^2+| w_{i,j}^{z} |^2 }$.
\STATE Compute HMT parameters $\theta_{i,j} = \{ \epsilon_{i,j}^{T},\sigma_{i,j}^{S},\sigma_{i,j}^{L} \}$.
\ENDFOR
\end{algorithmic}
\textbf{Output}: A set of feature vectors $\Theta = \big\{ \theta_{i,j} \big\}_{i,j=1}^{5,n_i}$.
\end{algorithm}

We divide paintings into patches of size $ 64 \times 64 $; features are extracted independently for each. 
HMT parameters are estimated by iterative Expectation Maximization (EM) method (see details in \cite{CrouseNB.ITSP1998}). The feature vector 
of each patch has $ 6\times  \left(2 \times 6+2 \times 4 \right) =120 $ entries: 2 variances for each of 6 scales
in each of 6 subbands, and 2 (non-redundant) transition probabilities for each of 
4 finest scales. (See Algorithm \ref{algo:featureextract}.) 

\section{Topic Modeling Based Style Analysis}
\label{sec:style}


The ``signature features'' making up the feature vectors can be viewed as primitive building blocks 
of the style of an artist; depending on the style of each artist, these signatures appear with different 
proportions \cite{JafarpourPBHBD.EUSIPCO2009,JohnsonHBBHDLPW.ISPM2008}. 
Usage patterns representing these proportions can be learned from the data. In a departure
from previous work, we propose to {\em learn} these usage patterns, 
using \emph{Latent Dirichlet Allocation (LDA)}, a probabilistic model for 
{\em discrete bag of words} data. The choice of stylistic elements in the 
creation of a painting is here viewed as similar to the choice of words 
in the process of creating a literary text. Learned topics, or stylistic patterns, 
are weights over the stylistic words; each painting is represented by weighted combinations 
of these patterns. This approach is similar to the bag of words model for object recognition 
in \cite{LiP.CVPR2005} which represents images with different proportions of object recognition 
feature patterns.

Since there are only five paintings, we divide each panel into sub-images with no overlap 
(these sub-images are larger than the small patches we used for feature extraction in 
Section \ref{sec:feature}). We consider each sub-image as an independent representation of the 
style of that panel's artist. We begin with definitions and show the mathematical formulation afterwards.
\begin{itemize}
\setlength{\itemsep}{0pt}%
 \setlength{\parskip}{0pt}
  \item The basic unit of a sub-image is a single \emph{patch} $\omega$, characterized by keywords 
indexed by $\{1,2,\dots,T\}$. (We will explain below how to generate these keywords.) A patch associated
to the $u$th keyword ($u=1,2,\dots,T$) is represented by a $T-$dimensional vector with all but the $u$th entry zero, 
and $\omega^{u}=1$. Two patches with the same keyword have the same drawing style.
  \item A \emph{sub-image} is composed of a sequence of $N$ patches (equivalent to one {\em article} in the LDA model in \cite{BleiNJ.JMLR2003}).
  \item An \emph{album} is a collection of sub-images, corresponding to a ``corpus'' in \cite{BleiNJ.JMLR2003}.
\end{itemize}

\algsetup{indent=0em}
\begin{algorithm}[t]
\caption{Keyword Labeling}
\label{algo:keywordlabel}
\textbf{Input}: All feature vectors $\Theta = \big\{ \theta_{i,j} \big\}_{i,j=1}^{5,n_i}$.\\
\textbf{Initialize}: Set $T=1$, $ \forall i,j, c_{i,j}^{1}=1$.
\vspace*{ 1 mm}

\textbf{Repeat} until $T=11$:

\begin{algorithmic}[1]
\FOR{$k=1:2^{T-1}$}
\STATE $ A_{T}^{k} = \bigcup_{i, j=1}^{5,n_i} \theta_{i,j}|_{c_{i,j}^{T}= k} $.
\STATE Divide $A_{T}^{k}$ into 2 subsets $A_{T+1}^{2k-1}$ and $A_{T+1}^{2k}$ using $K$-means algorithm, i.e., $A_{T}^{k} = A_{T+1}^{2k-1} \bigcup A_{T+1}^{2k}$ and $A_{T+1}^{2k-1} \bigcap A_{T+1}^{2k} = \emptyset$.
\STATE \emph{Label Update:} \textbf{If} $\theta_{i,j}|_{c_{i,j}^{T}= k} \in A_{T+1}^{2k-1}$, $c_{i,j}^{T+1}=2k-1$. \textbf{If} $\theta_{i,j}|_{c_{i,j}^{T}= k} \in A_{T+1}^{2k}$, then $c_{i,j}^{T+1}=2k$.
\ENDFOR
\end{algorithmic}
Set $T=T+1$.\\
\textbf{Output}: A set of labels $\big\{ c_{i,j}^{11} \big\}_{i,j=1}^{5,n_i} \in \{1, 2, \dots, 1024\}$.
\end{algorithm}

To generate the keywords, proceed as follows: cluster all patches with a divisive hierarchical 
clustering method, constituting a fixed vocabulary of $2^{10}=1024$ keywords in which patch $P_{i,j}$ 
is assigned a label $ c_{i,j}^{11} \in \{ 1,2,\dots,1024 \} $. This is a form of vector quantization; 
it represents each sub-image as a bag of words, see Algorithm \ref{algo:keywordlabel}. The binary structure 
of the labeling procedure ensures that labels that share dominant digit in their binary expansions are similar. 
``Stylistic patterns'' correspond to distribution over ``keywords''; our approach assumes that 
two sub-images having a similar distribution over keywords are similar in style. 

The Dirichlet priors that control each sub-image/pattern distribution and pattern/keyword distribution have
parameters $\bm{\alpha}$ and $\bm{\beta}$. These parameters are sampled once when generating a collection 
of images. If there is a collection of $K$ stylistic patterns, then the $K$-dimensional Dirichlet random 
variable $\bm{\pi}$, describing the pattern proportion for one sub-image, is sampled once per sub-image. The probability density function of $\bm{\pi}$ is given by
\begin{align}\label{eq:dirichlet}
p( \mbox{\boldmath $\pi$} | \mbox{\boldmath $\alpha$}) = \frac{\Gamma(\sum_{i=1}^{K} \alpha_i) }{\prod_{i=1}^{K} \Gamma(\alpha_i) } \pi_{1}^{\alpha_1-1} \dots \pi_{K}^{\alpha_K-1}
\end{align}

To generate a patch in a sub-image, the ``artistic process'' first chooses a stylistic pattern $t_{n}$ 
based on $\bm{\pi}$; this is a $K$-dimensional unit vector with $t_{n}^{j}=1$ if the $j$th pattern is selected. 
The pattern-keyword distribution $\bm{\phi}$ is a matrix of size $T \times K$, with
column $j$ representing the keyword distribution in pattern $j$. Then the process
produces a patch $\omega_{n} \sim p(\omega_{n} | t_n, \phi_j)$; note that $\phi_{l,j}=p( \omega_{n}^{l}=1 | t_{n}^{j}=1)$. Given $\bm{\alpha}$ and $\bm{\beta}$, the joint distribution of this generative model can be written as
\begin{align}\label{eq:lda}
\begin{split}
&p( \mbox{\boldmath $\pi$, t, $\omega$, $\phi$}  | \mbox{\boldmath $\alpha$, $\beta$} ) = \\
&p( \mbox{\boldmath $\pi$} | \mbox{\boldmath $\alpha$}) \prod_{j=1}^{K} p(\phi_j | \mbox{\boldmath $\beta$})  \prod_{n=1}^{N} p(t_{n} | \mbox{\boldmath $\pi$}) p(\omega_{n} | t_{n}, \phi_j)
\end{split}
\end{align}

\begin{figure}
\centering
\includegraphics[height=1in]{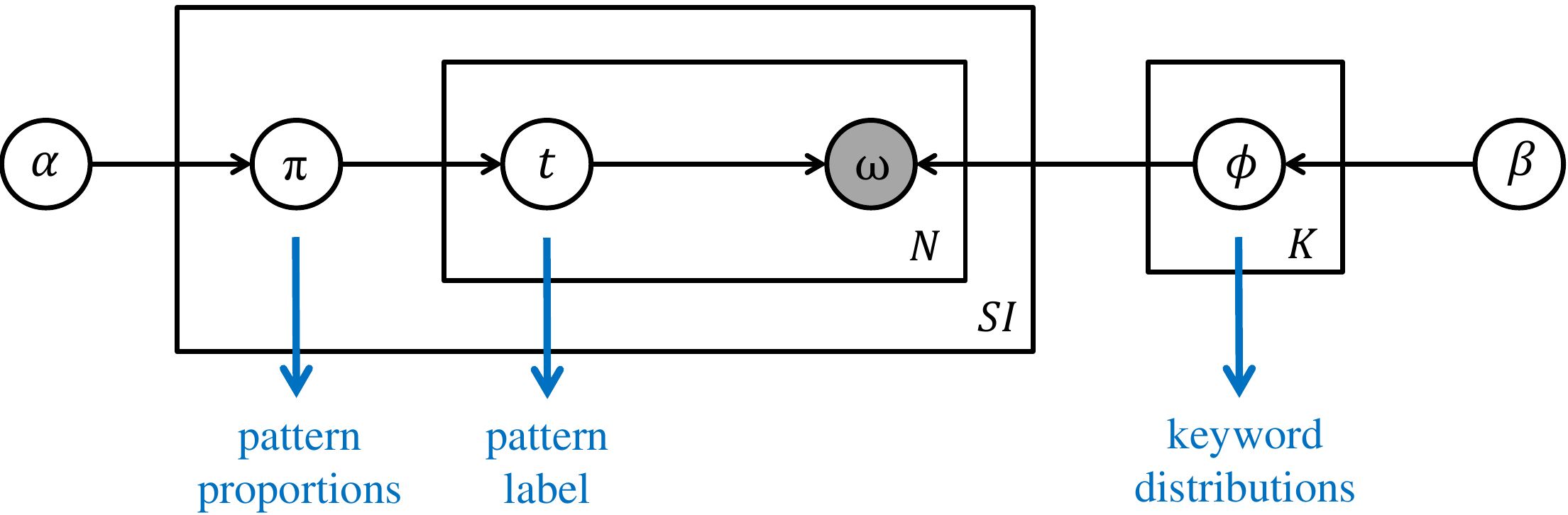}
\caption{Graphical model for the generative ``painting process''. Only one node is observed (shaded): the patches in sub-images. Boxes denote repeated processes. The different plates: $N$ is the collection of patches, $K$ is the stylistic patterns, $SI$ is the collection of sub-images.}\label{fig:ldaimage}
\end{figure}

A graphical description and dependencies of the generative model is shown in Figure \ref{fig:ldaimage}. The only observed variable is the patch, the hidden variables are sub-image/pattern distributions, patterns and pattern assignments. One needs to infer the hidden variables based on the observed variables, i.e., our goal is to compute $\bm{\pi}$ and $\bm{\phi}$ from $\bm{\omega}$. In \cite{BleiNJ.JMLR2003}, Blei et al. proposed an efficient variational EM algorithm for inference and parameter estimation; we use the \emph{Latent Dirichlet Allocation package} \cite{LdaPackage} for variational Bayesian (VB) implementation of LDA.

\section{Results and Discussion}
\label{sec:result}

We divide the scans of the 5 panels into 221 non-overlapping sub-images, 
each of size $480 \times 480$ pixels. In the feature extraction step we divide 
each sub-image into $64 \times 64$ small patches with 32 pixels overlap, so there are 
$ ((480-64)/32+1)^2=196 $ patches in one sub-image. Each patch is associated with one keyword
label; we observe the statistical keyword distribution
in each sub-image. The number of stylistic patterns, $K$, is set to be 20.

\begin{figure}
\centering
\includegraphics[height=2.25in]{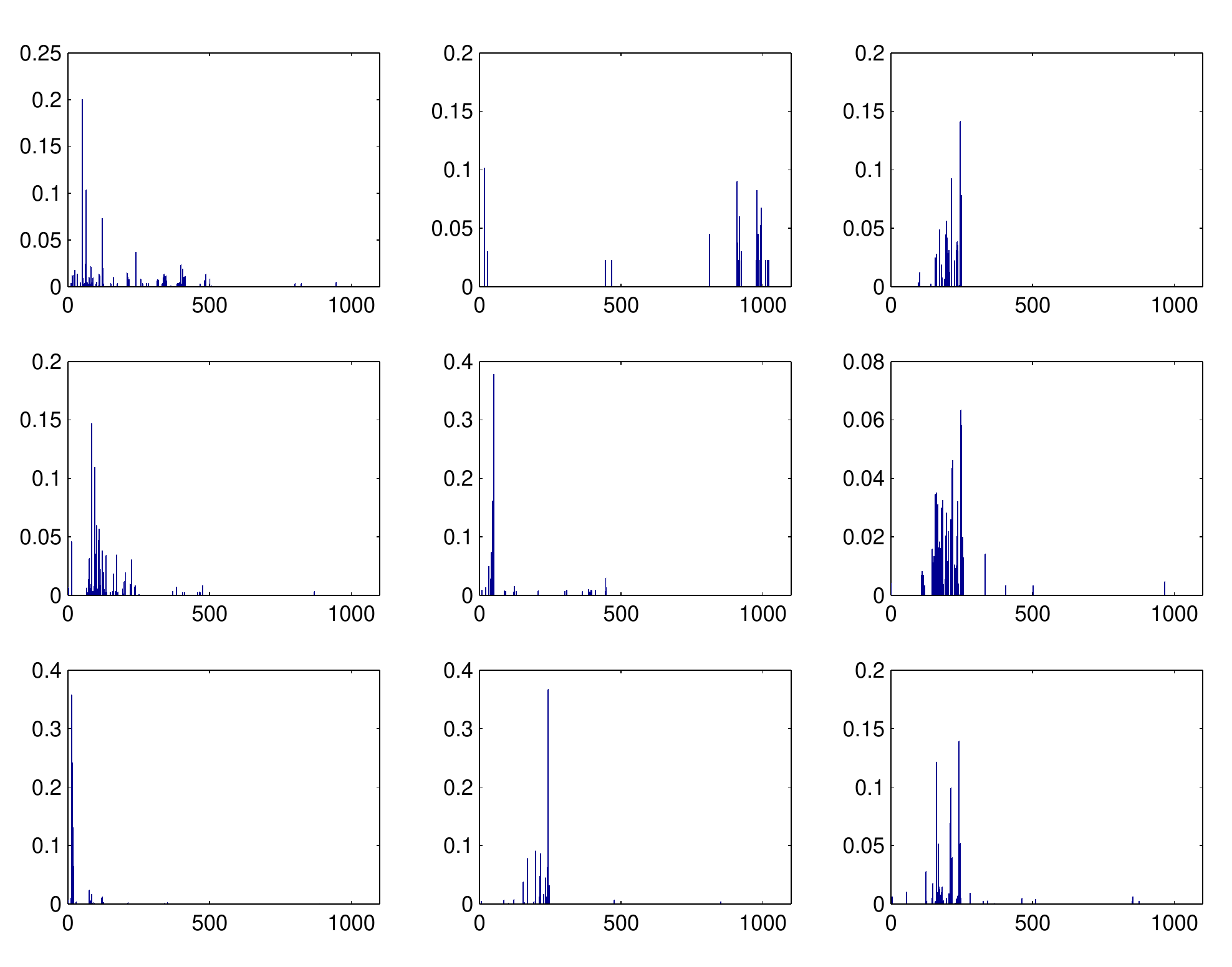}
\caption{ Keyword distribution of some stylistic patterns. \textbf{Left}: patterns 6, 8, 9. \textbf{Middle}: patterns 10, 12, 15. \textbf{Right}: patterns 16, 18, 20.}\label{fig:ldawordtopic}
\end{figure}

\begin{figure}
\centering
\includegraphics[height=2.45in]{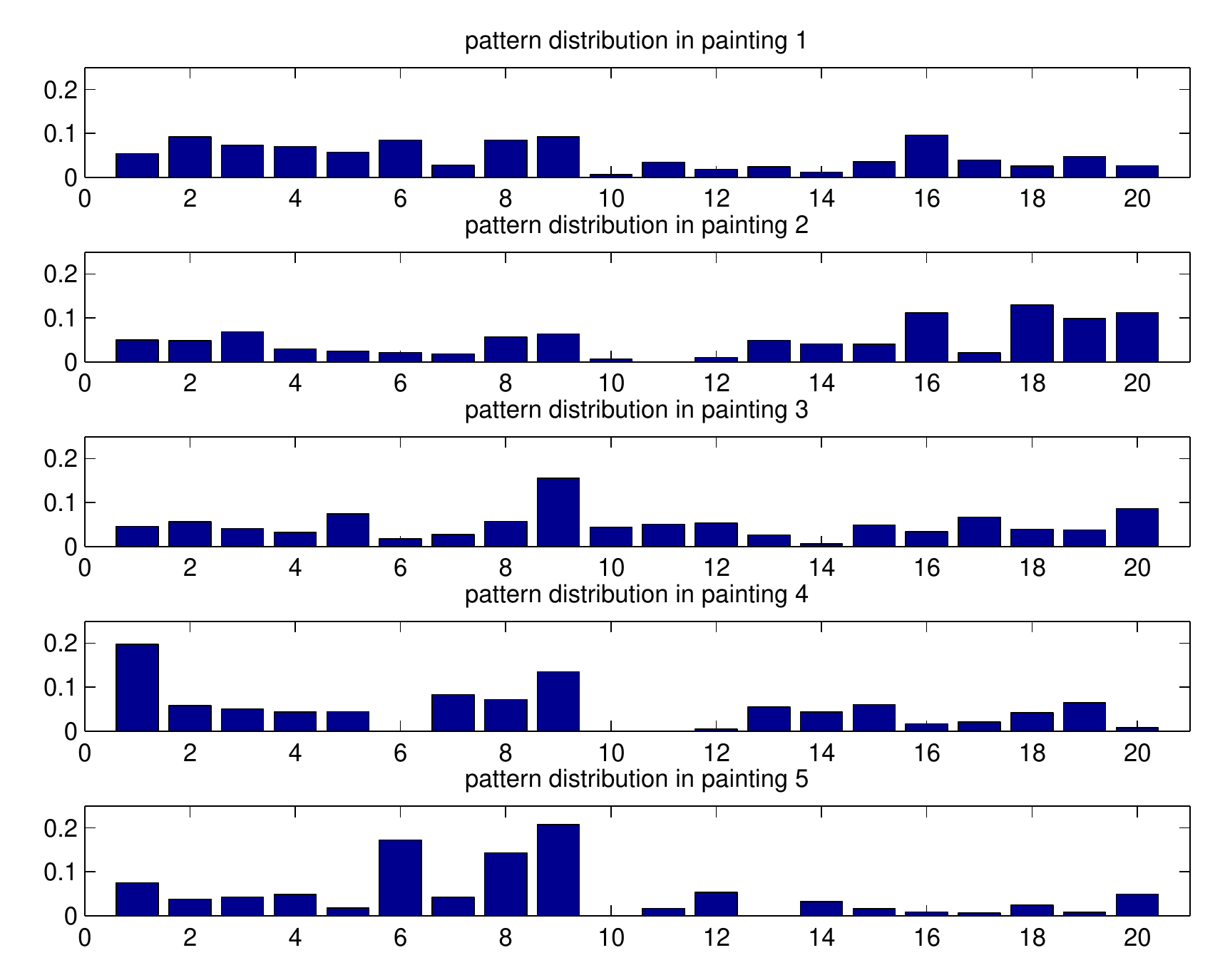}
\caption{ Stylistic pattern distributions in the five panels. For each painting, the x-axis denotes the pattern label; the y-axis corresponds to the total weight of each pattern in the painting.}\label{fig:ldatopicpaint}
\end{figure}

\begin{figure*}[t]
\subfigure[]{\includegraphics[width=2.65in]{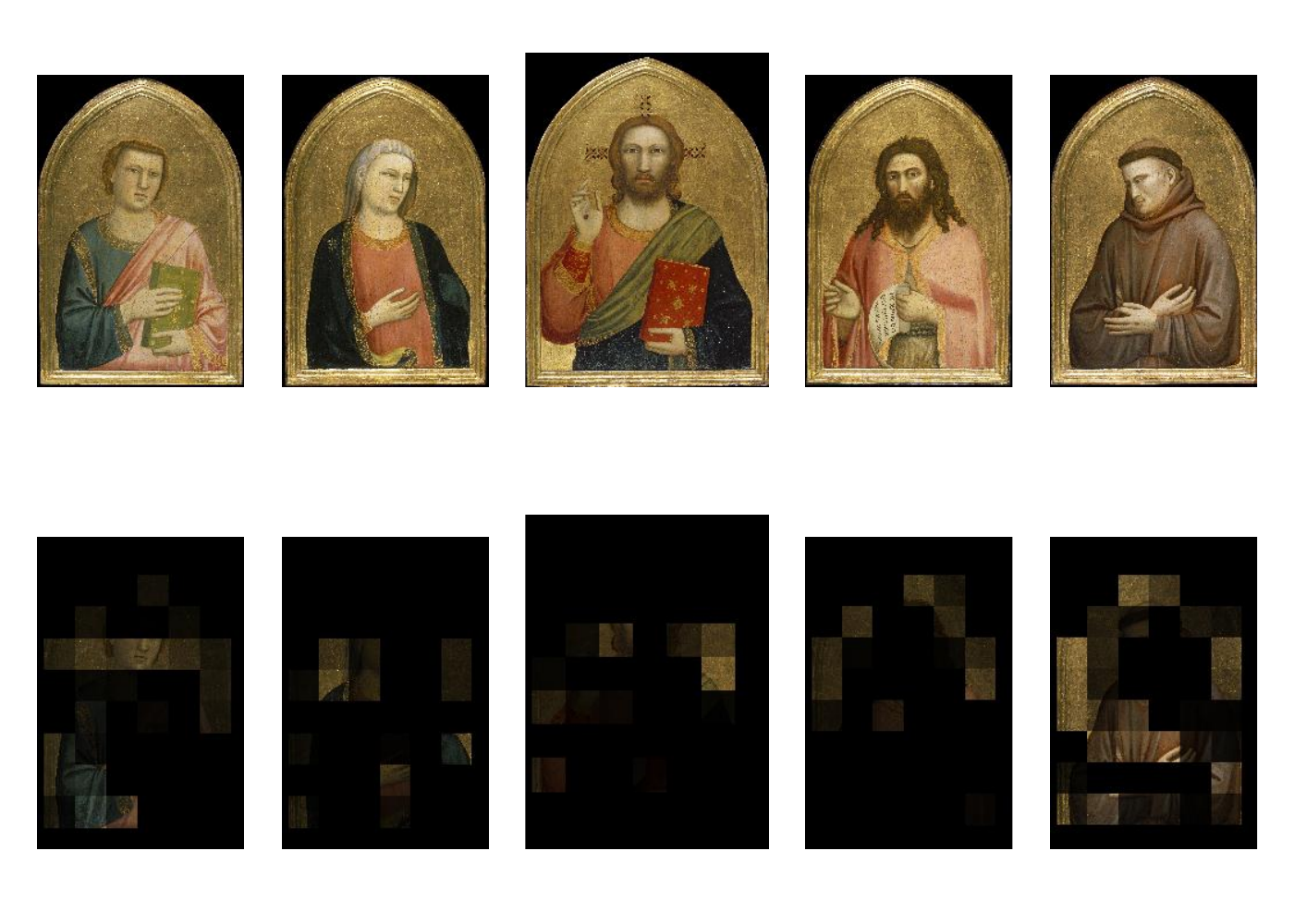}\label{fig:painting5}}
\hspace*{.03in}
\subfigure[]{\includegraphics[width=2.65in]{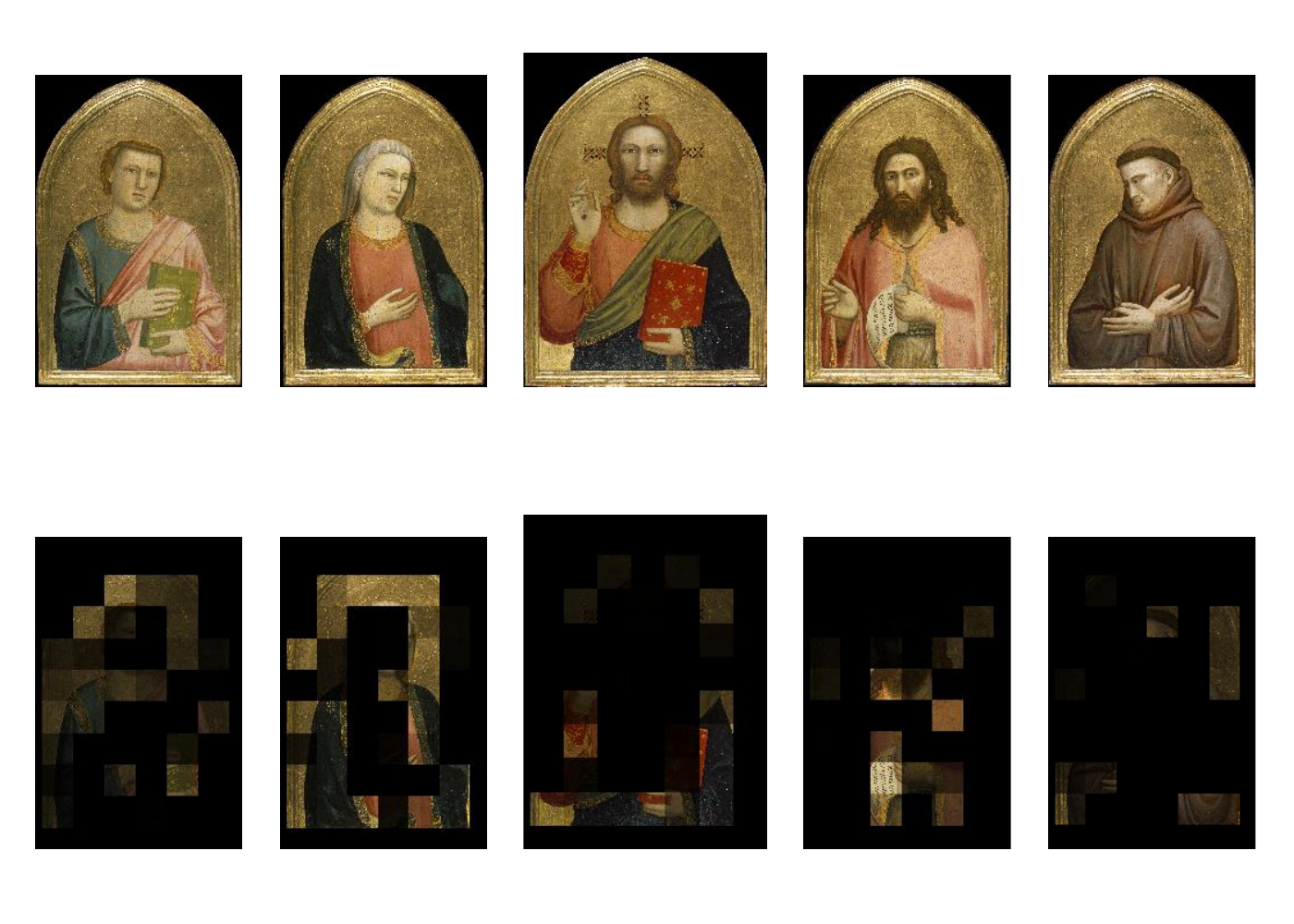}\label{fig:painting2}}
\hspace*{.03in}
\subfigure[]{\includegraphics[height=1.8in,width=1.5in]{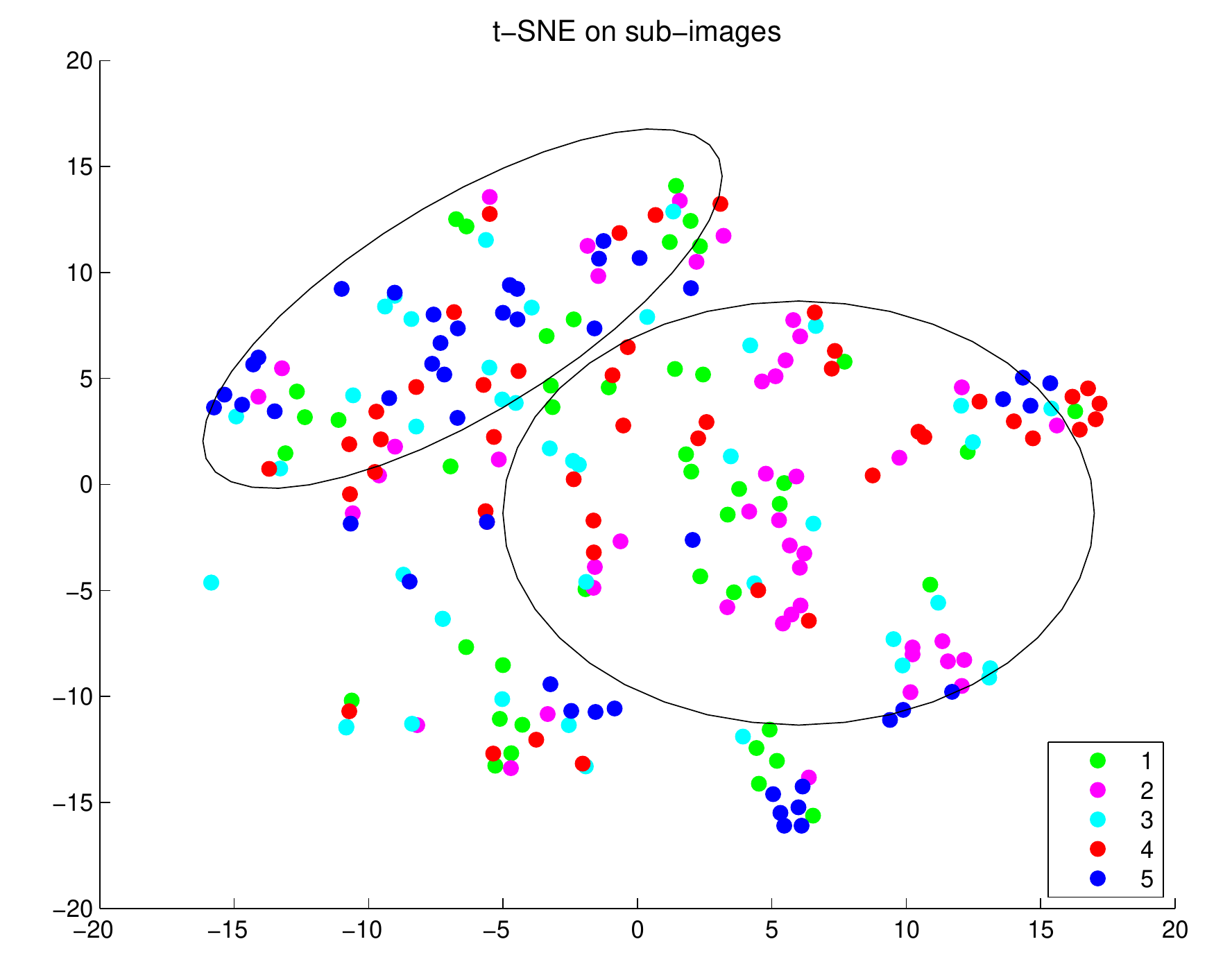}\label{fig:t-SNE}}
\caption{ (a) Illustration of the relative importance of patterns 6 and 8 in sub-images in the 5 panels. (b) Same for patterns 16, 18, 19 and 20. (c) Similarity of sub-images visualized via t-SNE; sub-images from the same painting are dots of the same color. }
\end{figure*}

We obtain two main parameters from LDA inference: the distribution function over the 
keywords for each pattern, and the weights over patterns in each sub-image. 
The keyword distributions for some patterns are in Figure \ref{fig:ldawordtopic}; each pattern is a 
sparse combination of the keywords in the vocabulary. Some patterns are highly weighted over keywords with
labels with the same dominant binary digits (for example, patterns 16 and 18 are both concentrated on 
keywords indexed from 128 to 255); these patterns are thus similar 
(see Section \ref{sec:style}) .

The pattern distribution 
for each of the five panels is obtained by adding the pattern weights of all 
its sub-images and normalizing the weights. Figure \ref{fig:ldatopicpaint} shows, e.g.,  that 
patterns 6 and 8 are more heavily represented in panel 5 than in the other four. 
Figure \ref{fig:painting5} visualizes this in terms of the original panels: each sub-image in each 
panel is shown with a brightness proportional to the sum of the weights of patterns 6 and 8 in that sub-image. 
Figure \ref{fig:painting2} does the same for patterns 16, 18, 19 and 20, which are clearly more heavily represented
in panel 2. 

Figure \ref{fig:t-SNE} illustrates the differences between the panels differently. 
Each sub-image is characterized by a 20-dimensional pattern distribution vector $\bm{\pi}$. 
We use the t-SNE algorithm of \cite{MaatenH.JLMR2008} to visualize all sub-images by projecting the pattern distribution matrix onto an optimal two-dimensional plane. Most of the blue dots (sub-images of panel 5) 
are located in one ellipse and a large percentage of the pink dots (panel 2) in another, disjoint ellipse; 
the sub-images in other three paintings are widely distributed in the plane. Once again panels 2 and 5 (the Virgin Mary and St. Francis) stand out.

\section{Conclusions}
\label{sec:con}

In this paper, we motivated and introduced a successful application of probabilistic topic modeling 
in painting analysis. 

The results are preliminary but promising. This project started when two of us (D. S. and W. B.) asked the 
other authors whether the techniques of \cite{WolffMJDC.ICASSP2011} could be used to quantify stylistic 
differences art historians perceive among the five panels of the Peruzzi Altarpiece. It is known that the panels were painted 
on wood from one single plank, that they all were prepared in the same way, that the underdrawings are similar; a recent study by the Getty Institute also showed similarities between pigments in the red, pink, blue, brown hair, and flesh paints. Yet the same
study also found differences making especially the panel depicting St. Francis (panel 5 in our numbering) stand out.
It is intriguing that this painting likewise stood out most in our analysis, as being ``stylistically'' more different. On the other hand, the Getty study found 
more reason to single out the panel of John the Baptist (panel 4) than the Virgin Mary (panel 2). 

Future work includes increasing the adaptivity of our model in the area of painting analysis, restricting it to subsets of the panels (only faces, or hands, or draped fabric, or hair). We intend to check, in particular, to what extent specific elements can be separated by our methods; concentrating on hair, for instance, the thin parallel brush work in the mustache/beard in Christ (panel 3) and John the Baptist (panel 4) seems, to the art historian's eye, possibly from a different hand than other parts of the altarpiece, and it would be interesting to see whether this is mirrored by a more adapted image analysis of the type described in this paper. Detailed images of the same five panels with other imaging modalities, such as X-ray radiography and infrared reflectography, are also available; we would like to fold the results of all these techniques together in a more ambitious, multispectral analysis. Finally, there have been several very interesting articles in the art history literature which attempt to establish similarities and differences in techniques between those paintings thought to be by Giotto and others in the workshop or followers; in this framework, it would be useful to extend our work to this larger body of paintings.

\vfill
\vfill

\bibliographystyle{IEEEbib}
\bibliography{refs}

\end{document}